\pdfoutput=1

\documentclass[11pt]{article}

\usepackage[]{acl}

\usepackage{lipsum}

\usepackage{times}
\usepackage{latexsym}

\usepackage{graphicx}
\usepackage{picinpar}
\usepackage{subcaption}
\usepackage{colortbl}
\usepackage{booktabs}
\usepackage{multirow}
\usepackage{soul}
\usepackage{enumerate}
\usepackage{color}
\usepackage{hyperref}
\usepackage{amsmath}
\usepackage{amssymb}
\usepackage{algorithm}
\usepackage{algpseudocode}

\usepackage{amsmath}

\usepackage[T1]{fontenc}

\usepackage[utf8]{inputenc}

\usepackage{microtype}

\title{Cross-Modality Gated Attention Fusion for Multimodal Sentiment Analysis}

\author{Ming Jiang  \\
Aalto University, Finland \\
  \texttt{ming.jiang@aalto.fi} \\
  \And
  Shaoxiong Ji \\
  Aalto University, Finland \\
\texttt{shaoxiong.ji@aalto.fi}\\
}

\begin{document}
\maketitle
\begin{abstract}
Multimodal sentiment analysis is an important research task to predict the sentiment score based on the different modality data from a specific opinion video. 
Many previous pieces of research have proved the significance of utilizing the shared and unique information across different modalities. 
However, the high-order combined signals from multimodal data would also help extract satisfied representations. 
In this paper, we propose CMGA, a Cross-Modality Gated Attention fusion model for MSA that tends to make adequate interaction across different modality pairs. 
CMGA also adds a forget gate to filter the noisy and redundant signals introduced in the interaction procedure. 
We experiment on two benchmark datasets in MSA, MOSI, and MOSEI, illustrating the performance of CMGA over several baseline models. 
We also conduct the ablation study to demonstrate the function of different components inside CMGA.
\end{abstract}

\section{Introduction}

With the vast popularity of social media in recent years, we can get a connection with more kinds of content, not only the text and figures but also the videos. 
Videos, including movies, TV shows, and short-form video streams, naturally consist of multimodal data types: the content of the text, visual (video frames), and acoustic (voice of speakers). 
Moreover, many of them usually have specific sentiment tendencies, which express the current emotional status of the speakers. 
It is essential to understand these sentiment tendencies. 

Across different modalities, there would be some shared and some unique information. 
Given an opinion video, its text is a pure statement of the opinion and contains semantic information. 
The visual modality exposes speakers' expressions.
The acoustic modality captures the tone of voice. 
These latter two types of information can not only be supplementary to understand the video better but also correct the misleading information in the textual modality \citep{ngiam2011multimodal}. 
As a result, the features contained in different modalities could help predict a more accurate sentiment score.

\begin{figure}[h!]
    \begin{center}
    \includegraphics[width=0.5\textwidth]{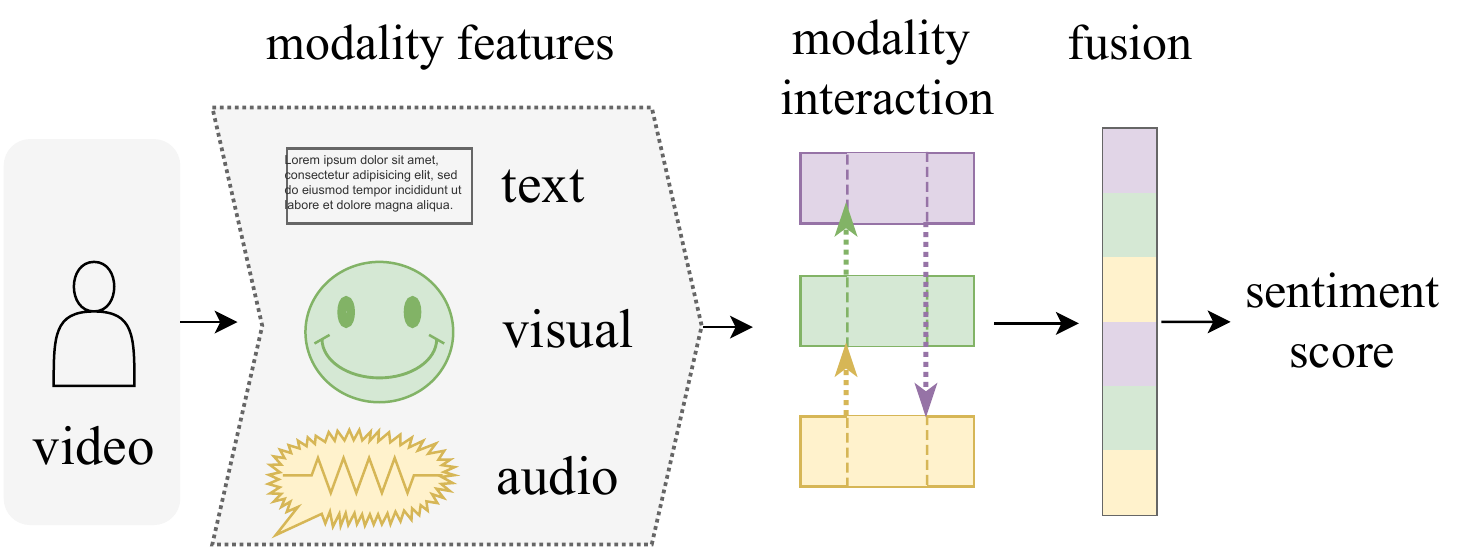}
    \caption{Multimodal Sentiment Prediction Pipeline. An opinion video includes three modalities: text, video, and audio. The interaction among modalities is useful for the prediction of an accurate sentiment score.}
    \label{fig:prob}
    \end{center}
\end{figure}

Many researchers have implemented deep learning methods on the multimodal data with good results achieved, illustrating the advantages of utilizing multimodal data beyond one single modality. 
As for the Multimodal Sentiment Analysis~(MSA) problem, it is crucial to fuse the shared information across different modalities and keep the unique information of a single modality. 
Some research methods focus on finding a vector space that contains these two kinds of signals simultaneously. 
For example, \citet{zadeh2017tensor} makes the three modalities interact with each other by extending the respective vector space into a shared tensor space with the 3-fold Cartesian product. 
However, it is difficult for these methods to distinguish the unique signals inside different modalities from the common interacted vector space. 
The extended common interacted vector space would contain the noisy and redundant signals, which are misleading and not valid for the downstream classification task.
Other methods aim to learn the two kinds of information separately. 
For example, \citet{MISA} proposes a model to learn two different subspaces of the three modalities, one of which is the shared encoder space. 
At the same time, the other one is the private space for maintaining unique signals. 
However, it lacks adequate interaction across the extracted information in subspaces of different modalities.

We propose a Cross-Modality Gated Attention~(CMGA) fusion model to alleviate these issues. 
CMGA aims to learn the cross-modality features that best summarize the interaction signals across different modalities and maintain the signals that are effective for MSA.
Specifically, we first divide the three modalities into three different pairs and generate the cross-modality attention feature maps of them, which are motivated by the design of the attention mechanism proposed by \citet{vaswani2017attention}. 
After this, CMGA pass the cross-attention maps into a forget gate designed to filter the noisy and redundant signals contributing little to the downstream prediction task. 
We use a residual connection proposed by \citet{he2016deep} to enhance the original modality signal and avoid the degradation problem. 
Finally, we input the cross-modality interaction features into a transformer-based fusion layer to predict sentiment scores. 
We evaluate the performance of CMGA on two famous benchmark datasets, CMU-MOSI and CMU-MOSEI, collected by \citet{zadeh2018multi}. 
Experiments show that CMGA outperforms several baseline models in most of the evaluation metrics. 
In addition, we also conduct two ablation studies on CMGA to illustrate the role of different components and the role of different modality data.

\section{Methodology}
\subsection{Problem Setup and Feature Extraction}
Given an utterance $U = \{u_{m}\}_{m \in \{t, v, a\}}$, its raw feature vectors contains three modalities in form of text, video and audio.
Our goal is to find the optimal interacted feature $h_{(t, v, a)}$ cross the three modalities that best represent the original utterance $U$ and predict the sentiment score $\Tilde{y}$ that close to the ground truth $y \in \mathbb{R}$.

We utilize pre-trained BERT~\citep{devlin2018bert} as the feature extractor for the textual inputs and obtain the final embedding of textual features by averaging the tokens' representations across the hidden states.
The visual and acoustic features are obtained by a stacked bi-directional Long Short Term Memory~(bi-LSTM)~\citep{hochreiter1997long}. 
Here, we get the embedding of the two modalities $u_v$ and $u_a$ by projecting the final state of LSTM into a fully-connected~(FC) layer.
We project each modality's features with a linear layer to obtain the representations of each modalities $\mathbf{z}_{m\in \{t, v, a\}} \in \mathbb{R} ^{d_k} $ with the same dimension size $d_k$.

\subsection{Modality Interaction}
\label{sec:interaction}
\subsubsection{Cross-modality Attention}
\label{sec:memory}

\begin{figure*}
\centering
    \includegraphics[width=0.8\textwidth]{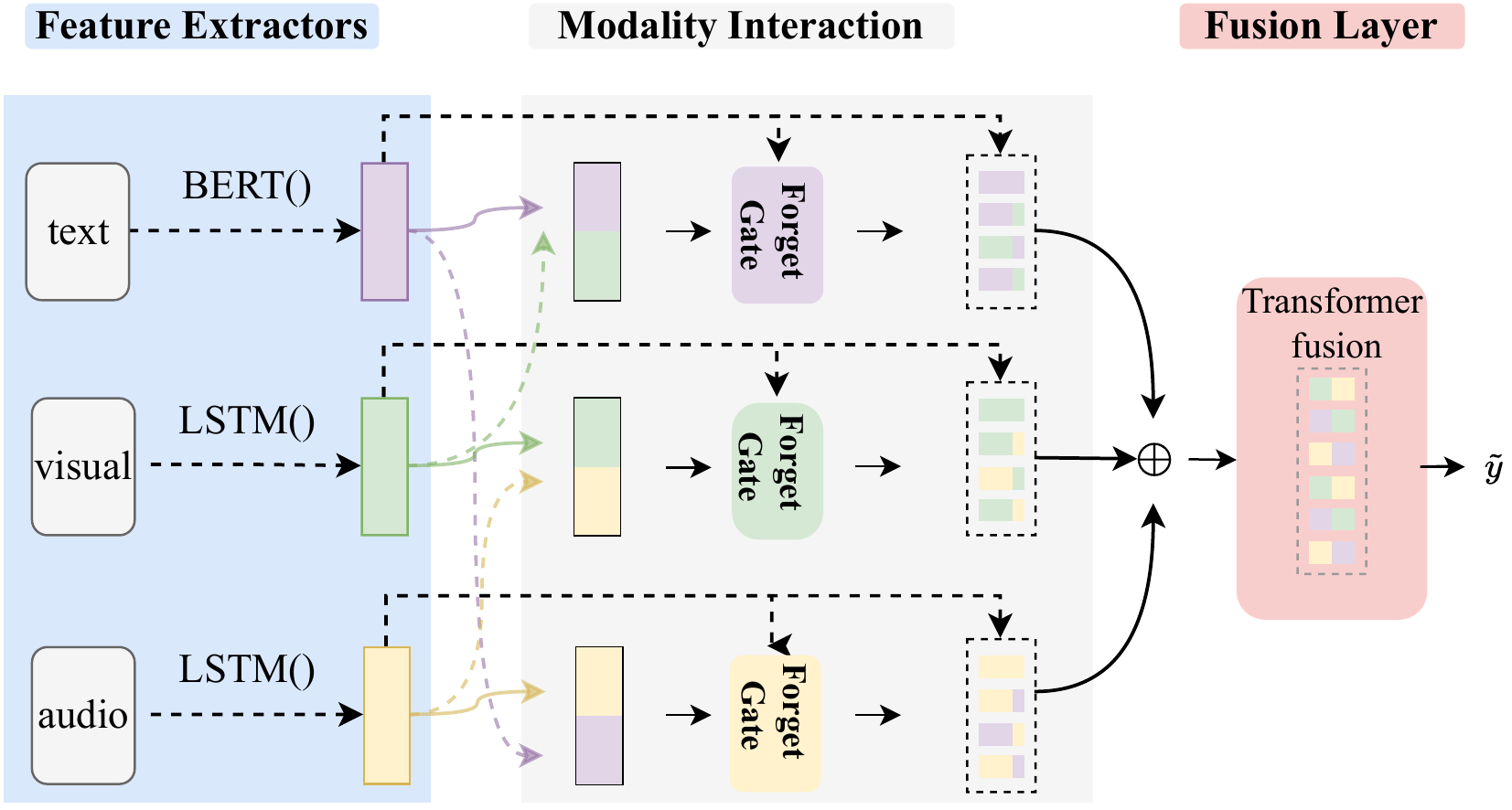}
    \caption{Architecture of CMGA. The multimodal features are separately extracted from respective models, and calculated the cross-modality interaction features within modality pairs. The fusion layer is a transformer encoder layer followed by a fully-connected layer.}
    \label{fig:framework}
\end{figure*}

The cross-modality attention aims to get the interacted signals across different modalities. 
Every two modalities would be one pair of inputs, i.e., \textbf{\textit{i)}} from text to visual;  \textbf{\textit{ii)}} from visual to acoustic; \textbf{\textit{iii)}} from text to acoustic. 
Here, we define the modality pair as a set $P = \{(t, v), (v, a), (t, a)\}$. 

The interaction feature generator inputs a pair of modalities $\mathbf{z_{(i, j), (i, j) \in P}}$.
The first one is utilized to generate \textit{key} and \textit{value} matrices, while the second modality is used to generate the \textit{query} matrix, i.e., $\mathbf{Q_{(i, j)}} = \mathbf{z_j} \mathbf{W^{Q_j}}$, $\mathbf{K_{(i, j)}} = \mathbf{z_i} \mathbf{W^{K_i}}$, and $\mathbf{V_{(i, j)}} = \mathbf{z_i}$.
Then, the cross-modality attention $\mathbf{a}_{(i, j)}$ of modalities pair~$(i, j)$ is transformed via a scaled dot-product~\citep{vaswani2017attention}. 
We obtain the cross-modality attention features $\mathbf{A} =\{\mathbf{a}_{(i, j)}\}_{(i, j) \in P}$ from each modality.

\subsubsection{Cross-modality Forget Gate}
\begin{figure}[h!]
    \begin{center}
    \includegraphics[width=0.49\textwidth]{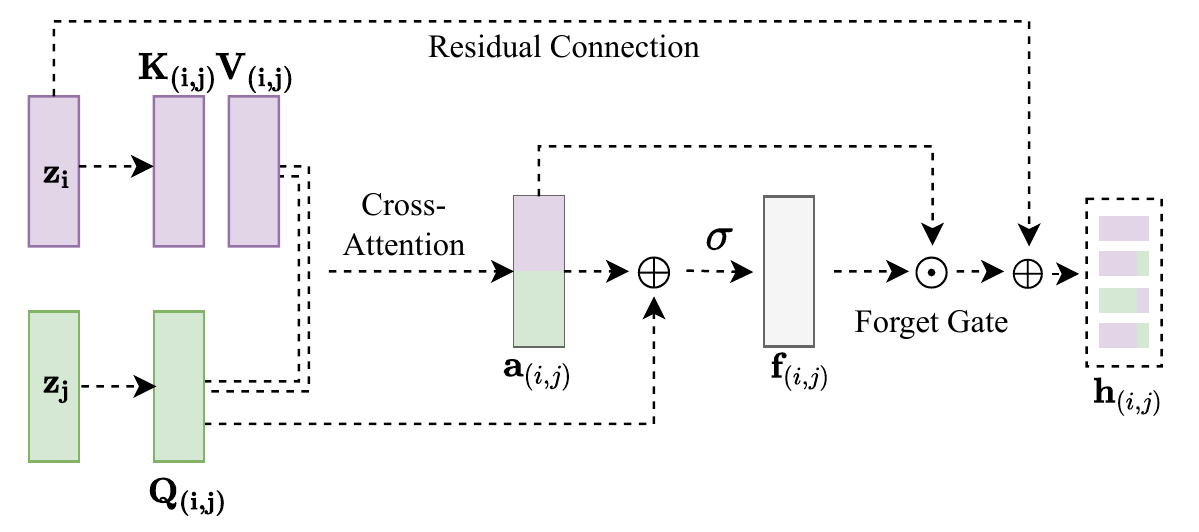}
    \caption{Cross-modality interaction architecture with one of the three modality pairs ($i, j$) for illustration}
    \label{fig:cross_mod}
    \end{center}
\end{figure}

The cross attention maps enables the model to capture the interaction between different modalities. 
However, $\mathbf{a}_{(i, j)}$ of a pair modality $(i, j)$ also contains plenty of redundant and noisy information. 
This part of information can obscure the instrumental interaction signals, leaving the original information of each modality still dominating the classification results in the downstream classification task. 
As a result, we might not fully exploit and utilize the additional interaction information across different modalities.
Therefore, motivated by the gated unit~\citep{cho2014properties}, we add a cross-modality forget gate to filter the redundant information and activate the useful cross-modality signal. 
As shown in Fig. \ref{fig:cross_mod}, the gate received the cross-modality attention generated in Section. \ref{sec:memory}, and pass it through a forget cell to generate the filtered cross-modality features. 
Specifically, the cross-modality attention map, which contains the modality pair's interacted signal, would first be used to generate a forget vector. 
The forget vector controls the information flow that would be memorized for the downstream classification tasks. 
The forget vector ${f}_{(i, j)}$ of modality pair $(i, j)$ is defined in Eq.~\ref{eq:forget_vector}.
\begin{equation}
\label{eq:forget_vector}
    \mathbf{f}_{(i, j)} = \sigma([\mathbf{a}_{(i, j)}\oplus \mathbf{z_j}] \mathbf{W^f} + \mathbf{b^f})
\end{equation}
Next, the filtered features of modality pair~$(i, j)$ are calculated as follows:
\begin{equation}
\label{eq:filtered}
    \mathbf{h}_{(i, j)} = \operatorname{ReLU}(\mathbf{z_i} + (\mathbf{a}_{(i, j)} \mathbf{W^m} + \mathbf{b^m}) \odot \mathbf{f}_{(i, j)}),
\end{equation}
where $\odot$ is the element-wise product between two vectors, $\oplus $ denotes the concatenation, and $\mathbf{W^f}$, $\mathbf{W^m}$, $\mathbf{b^f}$, $\mathbf{b^m}$ are trainable parameters. 
In addition, in Eq.~\ref{eq:filtered}, we keep $z_i$ to enhance the signal of original modality, which is motivated by the architecture of residual connection in ResNet proposed by \citet{he2016deep}.

\subsubsection{Fusion Layer}
The output of the forget gate consists of two parts: \textbf{\textit{i)}} the original modality feature; as well as \textbf{\textit{ii)}} the filtered cross-modality interaction. 
We obtain three pairs of outputs every two types of modalities: $\mathbf{ H = \{h_{i, j}\}_{i, j \in t, v, a}^{i\neq j} }$. 
In this part, we stack $H \in \mathbb{R} ^{3 \times d_h} $ across the three pairs and perform a multi-head self-attention on them to make each representation perceive the potential information remained in other cross-modality vector space, where $d_h$ is the dimension size of each cross-modality features through the forget gate. 
For each head $n$, we generate $\mathbf{Q_n} = \mathbf{H W^q_n}$, $\mathbf{K_n} = \mathbf{H W^k_n}$ and $\mathbf{V_n} = \mathbf{ H W^v_n}$, where $\mathbf{Q}, \mathbf{K}, \mathbf{V} \in \mathbb{R} ^{3 \times d_h}$ and $\mathbf{W^q_n}$, $\mathbf{W^k_n}$ and $\mathbf{W^v_n}$ are trainable parameters. 
Through the transformer fusion layer, we could obtain a new matrix $\Tilde{H}$ via a scaled dot-product attention in Eq.~\ref{eq:dot_product_attention}.
\begin{equation}
\label{eq:dot_product_attention}
    \Tilde{\mathbf{H_n}} = \operatorname{Softmax}(\frac{\mathbf{Q_n} \mathbf{K_n^\top}}{\sqrt{d_h}}) \mathbf{V_n} 
\end{equation}
The final output after fusion layer is $\Tilde{y} =  \Tilde{\mathbf{H}} \mathbf{W^o}$,
where $\Tilde{\mathbf{H}} = (\Tilde{\mathbf{H_1}}\oplus ... \oplus \Tilde{\mathbf{H_n}})$, and $\mathbf{W^o} \in \mathbb{R} ^{3 \times d_h \times 1}$ is trainable parameter.

\section{Experiments}

\begin{table*}[ht!]
\centering
\small
\setlength{\tabcolsep}{6pt}
\begin{tabular}{lccccc|ccccc}
\toprule
\multicolumn{1}{c}{\multirow{2}{*}{Models}} & \multicolumn{5}{c}{\textit{MOSI}} & \multicolumn{5}{c}{\textit{MOSEI}}                                                 \\
\multicolumn{1}{c}{}                        & \textit{MAE}   & \textit{corr}  & \textit{F-score} & \textit{Acc-2} & \textit{Acc-7} & \textit{MAE} & \textit{corr} & \textit{F-score} & \textit{Acc-2} & \textit{Acc-7} \\ 
\midrule
TFN	& 0.901          & 0.698          & 80.7             & 80.8           & 34.9           & 0.593        & 0.700         & 82.1             & 82.5           & 50.2           \\
LMF	& 0.917          & 0.695          & 82.4             & 82.5           & 33.2           & 0.623        & 0.677         & 82.1             & 82.0           & 48.0           \\
MFM	& 0.877          & 0.796          & 81.6             & 81.7           & 35.4           & 0.568        & 0.717         & 84.3             & 84.4           & 51.3           \\
ICCN	& 0.860          & 0.710          & 83.0             & 83.0           & 39.0         & 0.565        & 0.713         & 84.3             & 84.2           & 51.6           \\
MISA 	& 0.783          & 0.761         & 83.4             & 83.6           & 42.3          & 0.555        & 0.756         & 85.5             & 85.3   & 52.2 \\
\hline
TFN$^{\circ}$ 	& 0.893          & 0.705          & 79.6             & 80.3           & 35.6 & 0.609        & 0.712         & 81.9             & 82.3           & 50.0           \\
MISA$^{\circ}$	& 0.792          & 0.757          & 82.3             & 82.6           & 41.8          & 0.551        & 0.761         & 84.8             & 84.9           & 52.1           \\
\hline
CMGA	& {0.790} & {0.759} & {82.3}    & {82.7}  & {43.3} & {0.545				}    & {0.762}     & {85.0}        & {85.3}      & {53.0}      \\
\bottomrule
\end{tabular}
\caption{Performance comparison of baselines and CMGA on MOSI and MOSEI datasets. $^{\circ}$ means the performance of reproduced model. For those without the mark, the results are copied from the corresponding paper. The results of MISA are the best scores from the original paper.}
\label{tab:main_results}
\end{table*}

We experiment with two benchmarks with three modalities~(language, visual, and acoustic) in each utterance.
The CMU-MOSI dataset~\citep{zadeh2018multi} collects 2199 opinion video clips, each of which is annotated with a continuous sentiment score in the range of [-3, 3]. 
The value of the score represents the opinion attribution, where a smaller value close to -3 stands for negative sentiment and a larger value close to +3 is more positive.
The CMU-MOSEI dataset~\citep{zadeh2018multi} is an advanced version of the CMU-MOSI dataset, which collects more utterances from various speakers and topics. 
This dataset has 23,453 sentence utterance videos from more than 1000 online YouTube speakers with 250 different topics.

We compare the performance of CMGA with the following baseline models. 
\textbf{TFN}~\citep{zadeh2017tensor} utilizes the tensors' Cartesian space to calculate the multiplicative interactions between different modalities.  
\textbf{LMF}~\citep{LMF} uses a low-rank tensor to generate the representation of multimodal inputs efficiently.
\textbf{MFM}~\citep{MFM} introduces the joint generative-discriminative vector space, which factorizes representations into two sets of independent factors.
\textbf{MISA}~\citep{MISA} generates two independent vector subspaces to capture the shared and unique information of different modalities.
\textbf{ICCN}~\citep{ICCN} calculates the outer products between text, acoustic and text, visual features, and then implements a Canonical Correlation Analysis network for prediction.

In the LSTM models for acoustic and visual modalities, we implement a 2-layer bidirectional with 512-dimensional hidden states and layer norm between different layers. 
We implement a 12-layer transformer of 768-dimensional hidden states with 12 heads in the BERT model for textual modality. We use the pre-trained BERT tokenizers.
After extracting different modalities, we project them into the same size dimension of 128 by a fully-connected layer.
We use Mean Square Loss~(MSE) and Adam optimizer~\citep{kingma2014adam}. Our initial learning rate is 1e-4, and the model is trained on Tesla V100 GPUs. 

\subsection{Main Results}
Table \ref{tab:main_results} shows the predictive performance. 
CMGA outperforms all other models and has the most obvious improvement on \textit{acc-7} metric. 
CMGA outperforms MFM, MISA and ICCN, demonstrating the importance of learning the adequate interaction between different modalities.

\subsection{Analysis of Modalities and Neural Modules}
Table \ref{tab:ablation} shows the performance of CMGA without one specific modality. 
On both MOSI and MOSEI datasets, the textual modality $u_t$ plays the most important role. 
The performance drops sharply without $u_t$, showing that language conveys rich information for accurate prediction. 
We divide our modality interaction architecture into two separate parts, as described in Section \ref{sec:interaction}. 
Table \ref{tab:ablation} also shows the quantitative results of our model without one of the two components. 

\begin{table}[]
\small
\begin{tabular}{lllll}
\toprule
\multicolumn{1}{c}{\multirow{2}{*}{Models}} & \multicolumn{2}{c}{\textit{MOSI}}                   & \multicolumn{2}{c}{\textit{MOSEI}}                  \\
\multicolumn{1}{c}{}                        & \multicolumn{1}{c}{MAE} & \multicolumn{1}{c}{Acc-7} & \multicolumn{1}{c}{MAE} & \multicolumn{1}{c}{Acc-7} \\ 
\midrule
CMGA                                        & \textbf{0.790}          & \textbf{43.29}            & \textbf{0.545}          & \textbf{53.03}            \\
(-) text $u_t$                              & 1.591                   & 30.51                     & 0.818                   & 45.08                     \\
(-) video $u_v$                             & 0.804                   & 41.10                     & 0.547                   & 52.81                     \\
(-) audio $u_a$                             & 0.812                   & 42.10                     & 0.550                   & 52.79                     \\ \hline
(-) cross-attention                         & 0.845                   & 41.55                     & 0.587                   & 52.02                     \\
(-) forget gate                             & 0.856                   & 41.47                     & 0.594                   & 51.55                     \\
(+) bi-directional $h$                      & 0.792                   & 43.01                     & 0.550                   & 53.01                     \\ 
\bottomrule
\end{tabular}
\caption{Ablation study on the importance of modality and neural modules.~(-) represents missing for the mentioned factors, which include specific modality or model component.~(+) means add specific factors.}
\label{tab:ablation}
\end{table}

\subsection{Roles of Modality Interaction}
\label{sec:roleofmod}

We reverse each modality pair and check the performance of CMGA to further evaluate the order of modality pairs.
Table \ref{tab:order} shows that the order between video and audio does not affect the performance obviously, while the order of textual modality is critical. 
As illustrate in Section \ref{sec:memory}, the first modality $i$ in a pair $(i, j)$ is used to generate the key matrix $\mathbf{K_{(i, j)}}$ and value matrix $\mathbf{V_{(i, j)}}$. 
Inside the calculation of the attention map, we align the information of modality $j$ with modality $i$. 

\begin{table}[]
\small
\centering
\begin{tabular}{lcccc}
\toprule
\multicolumn{1}{c}{\multirow{2}{*}{Models}} & \multicolumn{2}{c}{\textit{MOSI}} & \multicolumn{2}{c}{\textit{MOSEI}} \\
\multicolumn{1}{c}{}                        & MAE             & Acc-7           & MAE              & Acc-7           \\ \midrule
CMGA                                        & 0.790           & 43.29           & 0.545            & 53.03           \\
$\sim$(text, video)                         & 0.814           & 41.22           & 0.561            & 52.13           \\
$\sim$(video, audio)                        & 0.791           & 43.27           & 0.547            & 53.01           \\
$\sim$(text, audio)                         & 0.804           & 42.27           & 0.551            & 52.84           \\ \bottomrule
\end{tabular}
\caption{Performance comparison with different orders of modality pairs. ($\sim$) represents to reverse the mentioned modality pair~($i, j$) into~($j, i$).}
\label{tab:order}
\end{table}

\section{Related Work}
\label{sec:related}
Sentiment analysis is a long-lasting research problem with many tasks such as aspect level sentiment analysis~\citep{lin2019deep}, emotion recognition in conversations~\citep{li2022bieru} and multimodal sentiment analysis~\citep{soleymani2017survey}. 
Our paper focuses on multimodal sentiment analysis.
This section reviews the modality interaction methods, which is trying to find the cross-modality features for different data modalities. 
Instead of the unimodal features, recent research has proved the significance of utilizing both the verbal and nonverbal information in multimodal sentiment analysis, such as the video and acoustic. 
\citet{zadeh2017tensor} proposed the tensor fusion network~(TFN) to obtain a cross-view feature by calculating a 3-fold Cartesian product. \citet{verma2020deep} implemented convolution calculation on the different cross-modality Cartesian spaces and fused them for the classification task. \citet{arevalo2017gated} placed a gated multimodal unit for modalities fusion. \citet{wang2020deep} utilized channel exchanging to make features of different modalities adequately integrated. \citet{yu2021learning} jointly training the multimodal and unimodal tasks, in which different modalities would be aligned in the unimodal tasks and interact with each other in the multimodal task.

Motivated by the success of transformers in many Natural Language Processing tasks, \citet{wang2020transmodality} proposed an end-to-end transformer-based model for sentiment analysis. This work proved the performance of transformer architecture in modality interaction. A transformer is built on the attention mechanism, whose intention to find the importance weights for different feature maps is suitable for the cross-fusion of different modalities. In addition, \citet{zadeh2018memory} implemented a Gated Memory Unit in the sequence learning to summarize the cross-view interactions learned through the attention units. 
In our work, the transformer-based model motivates the cross-modality interaction component. The attention mechanism idea proposed by \citet{vaswani2017attention} comes from the query system, which is suitable for different modalities to interact with each other in MSA tasks. In addition, different from the previous works, we aim to filter the noisy and redundant information that might be introduced in the modality interaction part.

\section{Conclusion}
This paper proposes CMGA, a multimodal learning framework that tends to predict sentiment scores by generating cross-modality interaction features. 
We combine the cross-attention map with the forget gate mechanism, which is helpful to get adequate interaction among different modality pairs and maintain the instrumental signals to represent the multimodal inputs. 
Our experiments show that CMGA achieves competitive predictive performance in most of the metrics. 
We evaluate the roles of importance for different modality features and the components inside the cross-modality interaction learning architecture, showing the importance of modality interaction.

\bibliography{multimodal_SA}

\begin{thebibliography}{22}
\expandafter\ifx\csname natexlab\endcsname\relax\def\natexlab#1{#1}\fi

\bibitem[{Arevalo et~al.(2017)Arevalo, Solorio, Montes-y G{\'o}mez, and
  Gonz{\'a}lez}]{arevalo2017gated}
John Arevalo, Thamar Solorio, Manuel Montes-y G{\'o}mez, and Fabio~A
  Gonz{\'a}lez. 2017.
\newblock Gated multimodal units for information fusion.
\newblock \emph{arXiv preprint arXiv:1702.01992}.

\bibitem[{Cho et~al.(2014)Cho, Van~Merri{\"e}nboer, Bahdanau, and
  Bengio}]{cho2014properties}
Kyunghyun Cho, Bart Van~Merri{\"e}nboer, Dzmitry Bahdanau, and Yoshua Bengio.
  2014.
\newblock On the properties of neural machine translation: Encoder-decoder
  approaches.
\newblock \emph{arXiv preprint arXiv:1409.1259}.

\bibitem[{Devlin et~al.(2018)Devlin, Chang, Lee, and
  Toutanova}]{devlin2018bert}
Jacob Devlin, Ming-Wei Chang, Kenton Lee, and Kristina Toutanova. 2018.
\newblock Bert: Pre-training of deep bidirectional transformers for language
  understanding.
\newblock \emph{arXiv preprint arXiv:1810.04805}.

\bibitem[{Hazarika et~al.(2020)Hazarika, Zimmermann, and Poria}]{MISA}
Devamanyu Hazarika, Roger Zimmermann, and Soujanya Poria. 2020.
\newblock Misa: Modality-invariant and-specific representations for multimodal
  sentiment analysis.
\newblock \emph{arXiv preprint arXiv:2005.03545}.

\bibitem[{He et~al.(2016)He, Zhang, Ren, and Sun}]{he2016deep}
Kaiming He, Xiangyu Zhang, Shaoqing Ren, and Jian Sun. 2016.
\newblock Deep residual learning for image recognition.
\newblock In \emph{Proceedings of the IEEE conference on computer vision and
  pattern recognition}, pages 770--778.

\bibitem[{Hochreiter and Schmidhuber(1997)}]{hochreiter1997long}
Sepp Hochreiter and J{\"u}rgen Schmidhuber. 1997.
\newblock Long short-term memory.
\newblock \emph{Neural computation}, 9(8):1735--1780.

\bibitem[{Kingma and Ba(2014)}]{kingma2014adam}
Diederik~P Kingma and Jimmy Ba. 2014.
\newblock Adam: A method for stochastic optimization.
\newblock \emph{arXiv preprint arXiv:1412.6980}.

\bibitem[{Li et~al.(2022)Li, Shao, Ji, and Cambria}]{li2022bieru}
Wei Li, Wei Shao, Shaoxiong Ji, and Erik Cambria. 2022.
\newblock {BiERU}: Bidirectional emotional recurrent unit for emotion detection
  in conversations.
\newblock \emph{Neurocomputing}, 467:73--82.

\bibitem[{Lin et~al.(2019)Lin, Yang, and Lai}]{lin2019deep}
Peiqin Lin, Meng Yang, and Jianhuang Lai. 2019.
\newblock Deep mask memory network with semantic dependency and context moment
  for aspect level sentiment classification.
\newblock In \emph{IJCAI}, pages 5088--5094.

\bibitem[{Liu et~al.(2018)Liu, Shen, Lakshminarasimhan, Liang, Bagher~Zadeh,
  and Morency}]{LMF}
Zhun Liu, Ying Shen, Varun~Bharadhwaj Lakshminarasimhan, Paul~Pu Liang, AmirAli
  Bagher~Zadeh, and Louis-Philippe Morency. 2018.
\newblock \href {https://doi.org/10.18653/v1/P18-1209} {Efficient low-rank
  multimodal fusion with modality-specific factors}.
\newblock In \emph{Proceedings of the 56th Annual Meeting of the Association
  for Computational Linguistics (Volume 1: Long Papers)}, pages 2247--2256,
  Melbourne, Australia. Association for Computational Linguistics.

\bibitem[{Ngiam et~al.(2011)Ngiam, Khosla, Kim, Nam, Lee, and
  Ng}]{ngiam2011multimodal}
Jiquan Ngiam, Aditya Khosla, Mingyu Kim, Juhan Nam, Honglak Lee, and Andrew~Y
  Ng. 2011.
\newblock Multimodal deep learning.
\newblock In \emph{ICML}.

\bibitem[{Soleymani et~al.(2017)Soleymani, Garcia, Jou, Schuller, Chang, and
  Pantic}]{soleymani2017survey}
Mohammad Soleymani, David Garcia, Brendan Jou, Bj{\"o}rn Schuller, Shih-Fu
  Chang, and Maja Pantic. 2017.
\newblock A survey of multimodal sentiment analysis.
\newblock \emph{Image and Vision Computing}, 65:3--14.

\bibitem[{Sun et~al.(2020)Sun, Sarma, Sethares, and Liang}]{ICCN}
Zhongkai Sun, Prathusha Sarma, William Sethares, and Yingyu Liang. 2020.
\newblock Learning relationships between text, audio, and video via deep
  canonical correlation for multimodal language analysis.
\newblock In \emph{Proceedings of the AAAI Conference on Artificial
  Intelligence}, volume~34, pages 8992--8999.

\bibitem[{Tsai et~al.(2018)Tsai, Liang, Zadeh, Morency, and
  Salakhutdinov}]{MFM}
Yao-Hung~Hubert Tsai, Paul~Pu Liang, Amir Zadeh, Louis-Philippe Morency, and
  Ruslan Salakhutdinov. 2018.
\newblock Learning factorized multimodal representations.
\newblock \emph{arXiv preprint arXiv:1806.06176}.

\bibitem[{Vaswani et~al.(2017)Vaswani, Shazeer, Parmar, Uszkoreit, Jones,
  Gomez, Kaiser, and Polosukhin}]{vaswani2017attention}
Ashish Vaswani, Noam Shazeer, Niki Parmar, Jakob Uszkoreit, Llion Jones,
  Aidan~N Gomez, {\L}ukasz Kaiser, and Illia Polosukhin. 2017.
\newblock Attention is all you need.
\newblock \emph{Advances in neural information processing systems}, 30.

\bibitem[{Verma et~al.(2020)Verma, Wang, Ge, Shen, Jin, Wang, Chen, and
  Liu}]{verma2020deep}
Sunny Verma, Jiwei Wang, Zhefeng Ge, Rujia Shen, Fan Jin, Yang Wang, Fang Chen,
  and Wei Liu. 2020.
\newblock Deep-hoseq: Deep higher order sequence fusion for multimodal
  sentiment analysis.
\newblock In \emph{2020 IEEE International Conference on Data Mining (ICDM)},
  pages 561--570. IEEE.

\bibitem[{Wang et~al.(2020{\natexlab{a}})Wang, Huang, Sun, Xu, Rong, and
  Huang}]{wang2020deep}
Yikai Wang, Wenbing Huang, Fuchun Sun, Tingyang Xu, Yu~Rong, and Junzhou Huang.
  2020{\natexlab{a}}.
\newblock Deep multimodal fusion by channel exchanging.
\newblock \emph{Advances in Neural Information Processing Systems},
  33:4835--4845.

\bibitem[{Wang et~al.(2020{\natexlab{b}})Wang, Wan, and
  Wan}]{wang2020transmodality}
Zilong Wang, Zhaohong Wan, and Xiaojun Wan. 2020{\natexlab{b}}.
\newblock Transmodality: An end2end fusion method with transformer for
  multimodal sentiment analysis.
\newblock In \emph{Proceedings of The Web Conference 2020}, pages 2514--2520.

\bibitem[{Yu et~al.(2021)Yu, Xu, Yuan, and Wu}]{yu2021learning}
Wenmeng Yu, Hua Xu, Ziqi Yuan, and Jiele Wu. 2021.
\newblock Learning modality-specific representations with self-supervised
  multi-task learning for multimodal sentiment analysis.
\newblock \emph{arXiv preprint arXiv:2102.04830}.

\bibitem[{Zadeh et~al.(2017)Zadeh, Chen, Poria, Cambria, and
  Morency}]{zadeh2017tensor}
Amir Zadeh, Minghai Chen, Soujanya Poria, Erik Cambria, and Louis-Philippe
  Morency. 2017.
\newblock Tensor fusion network for multimodal sentiment analysis.
\newblock \emph{arXiv preprint arXiv:1707.07250}.

\bibitem[{Zadeh et~al.(2018{\natexlab{a}})Zadeh, Liang, Mazumder, Poria,
  Cambria, and Morency}]{zadeh2018memory}
Amir Zadeh, Paul~Pu Liang, Navonil Mazumder, Soujanya Poria, Erik Cambria, and
  Louis-Philippe Morency. 2018{\natexlab{a}}.
\newblock Memory fusion network for multi-view sequential learning.
\newblock In \emph{Proceedings of the AAAI Conference on Artificial
  Intelligence}, volume~32.

\bibitem[{Zadeh et~al.(2018{\natexlab{b}})Zadeh, Liang, Poria, Vij, Cambria,
  and Morency}]{zadeh2018multi}
Amir Zadeh, Paul~Pu Liang, Soujanya Poria, Prateek Vij, Erik Cambria, and
  Louis-Philippe Morency. 2018{\natexlab{b}}.
\newblock Multi-attention recurrent network for human communication
  comprehension.
\newblock In \emph{Thirty-Second AAAI Conference on Artificial Intelligence}.

\end{thebibliography}
\bibliographystyle{acl_natbib}

\end{document}